%% file: aaai2026.tex
\documentclass[letterpaper]{article} 
\usepackage{aaai2026}  
\usepackage{amsmath}
\usepackage{amssymb}
\usepackage[dvipsnames]{xcolor}
\usepackage{times}  
\usepackage{helvet}  
\usepackage{courier}  
\usepackage[hyphens]{url}  
\usepackage{graphicx} 
\usepackage{natbib}  
\usepackage{caption} 
\usepackage{subcaption}
\usepackage{tabularx}
\usepackage{amsthm}
\newtheorem{proposition}{Proposition}
\usepackage{algorithm}
\usepackage{algorithmic}
\usepackage{placeins}

\usepackage{lipsum}

\usepackage{newfloat}
\usepackage{listings}
\DeclareCaptionStyle{ruled}{labelfont=normalfont,labelsep=colon,strut=off} 
\floatstyle{ruled}
\newfloat{listing}{tb}{lst}{}
\floatname{listing}{Listing}
\title{Relative Advantage Debiasing for Watch-Time Prediction in Short-Video Recommendation}
\author {
    Emily Liu, Kuan Han\thanks{Correspondence to kuan.han@bytedance.com}, Minfeng Zhan, Bocheng Zhao, Guanyu Mu, Yang Song
}
\affiliations {
    ByteDance Inc.\\
    \{emily.zfliu, kuan.han, zhanminfeng, zhaobocheng, muzhuoyu\}@bytedance.com, ys@sonyis.me
    
}

\usepackage{bibentry}

\begin{document}

\maketitle

\begin{abstract}
Watch time is widely used as a proxy for user satisfaction in video recommendation platforms. However, raw watch times are influenced by confounding factors such as video duration, popularity, and individual user behaviors, potentially distorting preference signals and resulting in biased recommendation models. We propose a novel relative advantage debiasing framework that corrects watch time by comparing it to empirically derived reference distributions conditioned on user and item groups. This approach yields a quantile-based preference signal and introduces a two-stage architecture that explicitly separates distribution estimation from preference learning. Additionally, we present distributional embeddings to efficiently parameterize watch-time quantiles without requiring online sampling or storage of historical data. Both offline and online experiments demonstrate significant improvements in recommendation accuracy and robustness compared to existing baseline methods.
\end{abstract}


\section{Introduction}
Immersive video viewing experience, such as TikTok and Reels, allows users to dive into a continuous stream of content that captures attention through full-screen visuals and intuitive swipe-based interactions.
It also brings unique challenges to recommender systems compared to traditional ones, as explicit interaction signals are often missing (e.g., clicks, ratings) or sparse (e.g., likes, comments, shares) \cite{xiaolin2023tpm, youtube}. This scarcity reduces the effectiveness of explicit feedback for learning user preferences. As a result, the duration a user spends viewing a video, commonly referred to as watch time, has become the standard implicit proxy for user interest. While watch time provides a continuous and behaviorally grounded measure of engagement, it is inherently confounded by factors unrelated to genuine preference \cite{D2Q}. For example, longer videos accumulate higher watch times regardless of true interest, leading to duration bias and a tendency to over-recommend long-form content \cite{D2Q, CWM}. Such biases can distort preference estimation and undermine recommendation quality \cite{WTG}. To overcome these limitations, robust debiasing strategies are needed to recover authentic user interests and ensure both fairness and effectiveness in recommendations \cite{Denoising_Implicit_Feedback_for_Recommendation}.
\newline \\
Prior work has primarily targeted duration bias by partitioning watch-time into length‐based buckets and applying transformations such as quantile normalization \cite{D2Q}, residual‐gain adjustment \cite{WTG, tang2023counterfactual}, and counterfactual or nonlinear mappings \cite{CWM}. While these techniques effectively mitigate video‐length distortions, they offer limited protection against other confounders such as content popularity or variations in user engagement patterns. More recent approaches model the full conditional watch‐time distribution by conditioning on each user–video pair to address variability in watch-time estimation \cite{CQE}, and may further incorporate additional context such as demographics and content category \cite{alignpxtr}. While this comprehensive approach captures uncertainty and heterogeneity, it suffers from the “single‐observation” problem: in practice, users rarely replay the same video, so there is almost never more than one sample under identical conditions. Consequently, distribution estimates derived from these methods tend to be noisy, prone to overfitting, and may introduce unintended dependencies between distribution fitting and recommendation goals, ultimately undermining preference modeling.
\newline \\
To address these challenges, we propose a relative advantage framework that debiases watch time by mapping each observed watch time onto two empirical reference distributions: one conditioned on video ID (aggregating all viewers’ watch times for that video) and the other on user ID (aggregating that user’s watch times across videos). Converting a watch time into its empirical quantile within each umbrella-conditioned distribution yields a uniform, bias-corrected signal that reflects relative engagement within each context. These umbrella factors can either be applied individually to correct video- and user-level confounders such as video length, popularity, and individual viewing habits, or combined through Bayesian evidence fusion to produce a robust and calibrated preference score. This design consists of two stage. The first stage involves estimating conditioned distributions; the second stage involves modeling preferences, ensuring modularity, numerical stability, and interpretability. Both offline benchmarks and live A/B tests demonstrate significant improvements in recommendation accuracy and robustness compared to existing baselines.
\newline \\
The major contributions of this work are:
\begin{itemize}
  \item A novel relative advantage debiasing framework for implicit watch-time feedback, correcting both item- and user-level confounders beyond simple duration bias.
  \item A two-phase architecture that explicitly decouples distribution estimation from preference modeling, enhancing training stability and model interpretability.
  \item A distributional embedding method to parameterize watch-time quantiles directly, eliminating the need for additional online indexing or historical data storage.
  \item Comprehensive offline and online evaluations demonstrating significant improvements in recommendation accuracy, robustness, and fairness over state-of-the-art baselines.
\end{itemize}

\begin{figure*}[ht]
\centering
    \includegraphics[width=0.98\linewidth]{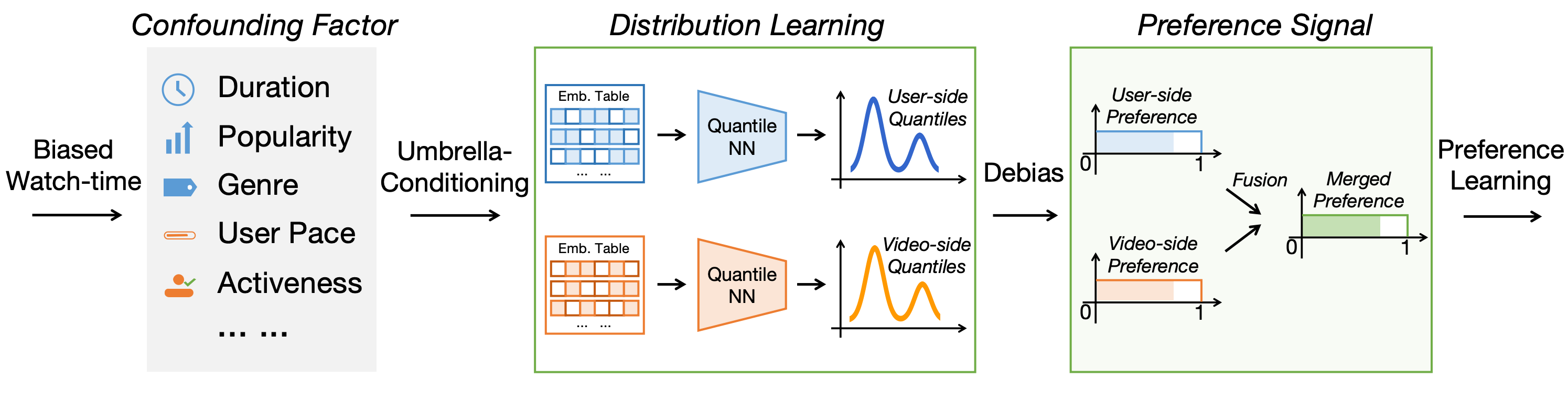}
        \caption{An overview of the RAD framework. First, distributional embeddings for user- and video-side confounders are learned from historical data. Then, these embeddings are used to measure the CDF of a biased watch-time value conditioned on user and video confounding variables. User-side and video-side signals are merged through Bayesian fusion to create a single bias-free label for preference learning.}
    \label{fig:structure}
\end{figure*}

\section{Related Work}
\paragraph{Watch-time Prediction}
Predicting video watch time is fundamental for video recommendation systems, as it serves as a primary proxy for user engagement. Early models, such as Weighted Logistic Regression (WLR) \cite{WLR} weight impressions by view duration but fail to handle the heavy-tailed and varying video length distributions, which are typical of short-form content. To better capture ordinal and uncertain aspects of watch time, Tree-based Progressive Regression (TPM) \cite{xiaolin2023tpm} discretizes watch time into ordered intervals and uses a hierarchical tree of binary classifiers. By contrast, CREAD \cite{cread} introduces error-adaptive bucketization to balance classification and restoration, effectively handling the long-tailed distributions commonly observed in streaming platforms.
\paragraph{Watch-time Debiasing}
A range of duration-debiasing methods have recently emerged. DVR~\cite{WTG} introduces Watch-Time-Gain, normalizing watch time within duration groups using adversarial learning. D2Q~\cite{D2Q} applies causal backdoor adjustments to control for duration effects, while CVRDD~\cite{tang2023counterfactual} uses counterfactual inference to directly eliminate duration bias at prediction time. Further developments also address measurement noise: D2Co~\cite{D2Co} simultaneously corrects for duration bias and noisy watching through GMM-based bias/noise estimation and correction, and CWM~\cite{CWM} introduces counterfactual watch time to recover engagement truncated by video duration. In contrast to these approaches, our method integrates both item-side and user-side relative advantage signals to jointly correct a range of biases, such as duration, popularity, and engagement patterns, within a single unified model, paving the way for a more generalizable and scalable debiasing framework.
\paragraph{Quantile Regression}
Quantile regression estimates conditional quantiles (such as the median or tail probabilities), providing a more comprehensive characterization of outcome distributions than summary statistics such as the conditional mean, especially in non-Gaussian or heavy-tailed data~\cite{koenker1978regression, meinshausen2006quantile, NBERw10428}. This technique has been effectively incorporated into machine learning models, including neural networks~\cite{nn_quantiles} and random forests~\cite{meinshausen2006quantile}, to yield uncertainty estimates and enhance robustness. In the context of video recommendation, D2Q \cite{D2Q} was among the first to leverage quantile regression within duration buckets to correct for duration bias in watch-time prediction. Conditional Quantile Estimation (CQE) \cite{CQE} extends this approach by predicting multiple watch-time quantiles per user–video tuple, enabling more flexible and context-aware modeling. AlignPxtr \cite{alignpxtr} further extends this paradigm by aligning predicted quantiles across various bias conditions (such as duration, demographics, or content category) through quantile mapping, further separating user interest from confounding effects. However, both CQE and AlignPxtr estimate conditional quantiles from only a single watch-time observation per user–video pair, which can lead to unstable and overfitted quantile estimates. In contrast, our approach aggregates watch-time distributions on both the item and user sides, enabling more robust and generalizable relative-advantage metrics. This separation of distribution estimation from preference modeling allows us to capture uncertainty and correct for multiple biases within a unified framework.

\section{Method}

\subsection{A Generative Model of Watch Time}
\label{sec:generative-model-watch-time}
We treat each observed watch time $S_{u, i}$ as a generative model with two sources: the true underlying preference user $u$ has for video $i$, and any systematic confounding factors outside of this preference, formally as
\begin{equation}
S_{u,i} = C_{u,i} + P_{u,i} + \varepsilon_{u,i},
\end{equation}
where $C_{u,i}$ captures the aggregate effect of confounders, $P_{u,i}$ reflects the genuine user--video preference, and $\varepsilon_{u,i}$ denotes irreducible noise.

To make the impact of confounding explicit, let $\mathcal{C}$ represent the collection (or $\sigma$-algebra) generated by all fine-grained confounders. Applying the law of total variance, the variability in observed watch time decomposes as
\begin{equation}
\mathrm{Var}(S_{u,i}) =
\underbrace{\mathrm{Var}\left( \mathbb{E}[S_{u,i} \mid \mathcal{C}] \right)}_{\text{bias variance}}
+ \underbrace{\mathbb{E}\left[ \mathrm{Var}(S_{u,i} \mid \mathcal{C}) \right]}_{\text{residual variance}}.
\end{equation}
Here, $\mathrm{Var}( \mathbb{E}[S_{u,i} \mid \mathcal{C}] )$ quantifies the bias variance, i.e., systematic differences in watch time induced by confounders, while $\mathbb{E}[ \mathrm{Var}(S_{u,i} \mid \mathcal{C}) ]$ is the residual variance attributable to true preferences or irreducible noise. By conditioning on confounders, we seek to minimize the bias variance, enabling the model to focus on the residual variation that more accurately reflects genuine user–video preferences and results in more stable, interpretable preference learning.

\subsection{Umbrella Conditioning for Variance Reduction}
In large-scale recommendation systems, the sheer number of confounders makes it impractical to correct for each one individually. However, many real-world confounders are deterministic functions of higher-level identifiers--what we refer to as \emph{umbrella factors}. For instance, every video-side confounder--such as duration, category, creator, or popularity--denoted $c^{(1)}, \ldots, c^{(K)}$, is determined by the video ID $i$. For $K$ confounders dependent on video ID, let $\sigma(c^{(1)}, \ldots, c^{(K)})$ be the $\sigma$-algebra generated by these fine-grained confounders, and $\sigma(i)$ that generated by video ID. Since
\begin{equation}
  \sigma\bigl(c^{(1)},\ldots,c^{(K)}\bigr)
  \;\subseteq\;
  \sigma(i),
\end{equation}
we have the following variance reduction property.

\begin{proposition}[Variance monotonicity under umbrella sufficiency]\label{prop:variance-monotonicity}
For any video-side confounder $c^{(k)}$,
\begin{equation}
  \mathrm{Var}\bigl(\mathbb{E}[S_{u,i}\mid i]\bigr)
  \;\ge\;
  \mathrm{Var}\bigl(\mathbb{E}[S_{u,i}\mid c^{(k)}]\bigr).
\end{equation}
\end{proposition}
Proof of this statement is provided in the Appendix. Intuitively, conditioning on video ID collapses confounders—such as duration, category, creator, and popularity—into a single context; the same logic applies to user ID, which subsumes factors like activeness level, device, and viewing habits. While any conditioning variable reduces bias variance to some extent, the amount removed depends on how much between-group variation it captures. Since user and video IDs encompass a broad range of real-world confounders, umbrella conditioning offers a scalable and effective way to produce cleaner residual signals for robust preference modeling.

\subsection{Conditional Quantile Transformation}
To achieve robust debiasing, we transform each observed watch time by its conditional cumulative distribution function (CDF), conditioned on an umbrella factor such as video or user ID. Let $F_{S|G}(s) = \Pr(S \leq s \mid G)$ denote the conditional CDF for umbrella factor $G \in \{i, u\}$. The within-context quantile is then defined as
\begin{equation}
    Q_{u,i}(G) = F_{S|G}(S_{u,i}) \in (0, 1).
\end{equation}
\begin{proposition}[Statistics and Independence of Quantile Labels]\label{prop:quantile_uniformity}
For any umbrella factor $G$, the quantile label $Q_{u,i}(G)$ has mean $\mathbb{E}[Q_{u,i}(G)] = \frac{1}{2}$ and variance $\mathrm{Var}(Q_{u,i}(G)) = \frac{1}{12}$, and is statistically independent of $G$, i.e., $Q_{u,i}(G) \perp\!\!\!\perp G$.
\end{proposition}
A proof of Proposition~\ref{prop:quantile_uniformity} is provided in the Appendix. This mapping ensures that quantile labels are bias-free, bounded, and homoskedastic, stabilizing both the quantile signal and optimization process. Notably, after this transformation, $Q(G)$ is statistically independent of confounders included in $G$—that is, all systematic group-level biases are removed, and the remaining variation reflects latent user--video preference. These properties make quantile (CDF) labels an effective and principled choice for preference modeling over alternatives like z-scores.

\paragraph{User-Side Duration Heterogeneity}
Unlike video ID, which encapsulates duration as an intrinsic attribute, user ID cannot fully account for duration—an especially dominant source of bias in watch-time ~\cite{D2Q}. To address this, we first split all watch times into $D$ near-equal-mass bins (e.g., $D=4$) based on duration~\cite{D2Q}. For each bin, we estimate the empirical user-side CDF $F_{S|u,\mathrm{bin}}(s)$, and assign the bin-specific quantile label to each watch time:
\begin{equation}
    Q_{u,i}^{(\mathrm{user},\mathrm{bin})} = F_{S|u,\mathrm{bin}}(S_{u,i}) \in (0,1).
\end{equation}
This additional conditioning ensures that user-side RAD labels remain robust to heterogeneity in video length—a correction not required for video ID umbrella as duration is already captured.

\subsection{Relative Advantage Debiasing (RAD)}
\label{sec:rad}
Based on the conditional quantile transformation introduced above, we propose \textbf{Relative Advantage Debiasing (RAD)}, a two-stage procedure leveraging hierarchical umbrella conditioning. RAD transforms watch times into quantile-based labels, providing stable, bounded, and unbiased signals of relative user engagement.

\paragraph{Stage 1: RAD Label Estimation.}
For each umbrella factor (video or user), we estimate empirical watch-time distributions from historical logs and derive quantile labels:
\begin{itemize}
\item  \textbf{Video-side RAD (RAD-V):} For each video $i$, we compute its empirical CDF, $F_{S|i}(s)$, as quantile labels:
\begin{equation}
Q_{u,i}^{(\mathrm{video})}=F_{S|i}(S_{u,i})=\Pr_{u'}(S_{u',i}\le S_{u,i}\mid i).
\end{equation}
\item \textbf{User-side RAD (RAD-U):} We partition watch times into $D$ bins, and compute the empirical CDF specific to the \(d\)-th bin as quantile labels:
\begin{equation}
Q_{u,i}^{(\mathrm{user})}=F_{S|u,\mathrm{d}}(S_{u,i})=\Pr_{i'}(S_{u,i'}\le S_{u,i}\mid u,\mathrm{d}).
\end{equation}
\end{itemize}
\paragraph{Stage 2: Preference Modeling.}
Given these debiased RAD labels, we train a parametric model (e.g., MLP, DCN, etc) to predict RAD-U or RAD-V label. Separating two stages yields disentangled, stable targets to learn. 

We evaluate RAD with two hypotheses. First, the bounded and homoskedastic nature of RAD labels makes training more stable and efficient, leading to lower watch-time prediction error (e.g., reduced MAE) when mapped back to raw values. We validate this by comparing RAD-based models with standard baselines on watch-time prediction. Second, by removing systematic biases, RAD labels offer a cleaner measure of user engagement, which should yield better ranking performance—higher XAUC/XGAUC and improved online metrics. We confirm this by ranking with RAD predictions and measuring gains in both offline and online evaluation.

\subsection{Dual‑sided Bayesian Evidence Fusion}
\label{sec:combined_cdf}
While RAD-U and RAD-V each correct user- and video-side biases, fusing them yields a unified, robust preference score. We achieve this by first mapping each quantile label into z-score space using the probit (inverse normal CDF) transform:
\begin{equation}
z_{u,i}^{(\mathrm{user})}
= \Phi^{-1}\bigl(Q_{u,i}^{(\mathrm{user})}\bigr),
\quad
z_{u,i}^{(\mathrm{video})}
= \Phi^{-1}\bigl(Q_{u,i}^{(\mathrm{video})}\bigr).
\end{equation}
We then combine the two z-scores through a weighted average, and mapping the fused score back to quantile space::
\begin{equation}
\widehat z_{u,i}
= \frac{\alpha\,z_{u,i}^{(\mathrm{user})} \;+\;\beta\,z_{u,i}^{(\mathrm{video})}}
       {\sqrt{\alpha^2 + \beta^2}}
\;\to\; Q_{u,i} = \Phi\bigl(\widehat z_{u,i}\bigr),
\end{equation}
where $\alpha$ and $\beta$ are precision weights that reflect the reliability of each signal. In offline experiments, we set these weights proportional to the statistical support, that is, the number of samples used to estimate the user-- and video--side CDFs. In large-scale or online deployment, we use equal weights ($\alpha = \beta = 1$), since both user and video CDFs are well-supported with sufficient data and this simplifies deployment.

This Bayesian fusion produces a stable, calibrated, and interpretable score that jointly corrects user- and video-side biases, naturally weighting each view by its confidence. The unified label integrates directly into downstream models and improves robustness, especially in cold-start scenarios where one side has limited data. An overview of the two-stage structure is given in Figure ~\ref{fig:structure}. The effect of the ratio $\beta/\alpha$ on results, and the robustness of RAD to distributional shift, is discussed in the Appendix. 

\section{Experiments}
We show that Relative Advantage Debiasing (RAD) significantly improves both watch-time prediction accuracy and ranking quality over state-of-the-art baselines. Further analysis confirms the effectiveness of RAD’s two-stage learning and Bayesian fusion, and highlights an efficient strategy for distribution learning. Finally, large-scale online A/B tests demonstrate clear gains in user engagement, underscoring RAD’s real-world impact.
\subsection{Dataset and Experiment Setup}

\subsubsection{Datasets}

We conduct offline experiments on two large-scale short-video datasets:
\textbf{(1) KuaiRand-Pure:} A public benchmark from the KuaiShou platform, widely used for debiasing tasks in sequential video recommendation. We use the KuaiRand-Pure subset as recommended by prior work~\cite{KR}, chronologically splitting data using a sliding timestamp cut-off, allocating training (79.6\%), validation (8.7\%), and test (11.6\%) sets. Users in validation or test sets are filtered to ensure they also appear in training. The final split contains 26{,}592 users, 7{,}146 items, and 1,384,425 user–video interactions.
\textbf{(2) An Offline Industrial Dataset:} We further evaluate RAD on a offline industrial dataset sampled from a large-scale short-video platform. The snapshot contains well over a billion user–video interactions, making it orders of magnitude larger than KuaiRand. Data are split chronologically: the first 14 days form the training set, and the following day is used for testing. This large-scale setting allows us to assess whether RAD’s improvements persist under high-data conditions.
\subsubsection{Baselines} 
We compare our method against a comprehensive set of state-of-the-art baselines covering direct regression, normalization, noise correction, and quantile-based modeling. Baselines include: Value Regression (VR), Play Completion Rate (PCR) \cite{CWM}, Weighted Logistic Regression (WLR) \cite{WLR}, Watch-time Gain (WTG) \cite{WTG}, Debiased and Denoised Correction (D2Co)\cite{D2Co}, Duration-debiased Quantiles (D2Q) \cite{D2Q}, Counterfactual Watch Model (CWM) \cite{CWM}, and Conditional Quantile Estimation (CQE) \cite{CQE}. Detailed descriptions of baselines are given in the appendix, while implementations follow the corresponding papers and established protocols~\cite{CWM,CQE}.
\subsubsection{Backbones and Hyperparameters} 
All methods are implemented with the following backbone architectures to test generalizability, including multi-layer perceptron (MLP), the state-of-the-art Deep \& Cross Network (DCN) \cite{DCN}, and its recent extensions, DCN\_v2 \cite{DCNV2} and GDCN \cite{GDCN}. A description of hyperparameters is given in the Appendix.
\subsubsection{Evaluation Metrics}

\setlength{\dbltextfloatsep}{6pt}    
\setlength{\belowcaptionskip}{4pt}   
\begin{table*}[ht]
    \centering
    
\setlength{\tabcolsep}{4pt}
\renewcommand{\arraystretch}{0.95}
    \begin{tabular}{c|c|cccccccc|cc|c}
    \hline
        Metric & Backbone & VR & PCR & WLR & WTG &D2Co & D2Q &CWM & CQE&RAD-V&RAD-U&RAD-UV \\
        \hline
        &MLP & 21.525 &45.912	&21.229	&22.627	&22.123	&19.763	&20.330	&21.434	&\textbf{18.050}	&\textbf{18.221}	&\textbf{18.050}\\
        MAE & DCN & 
        23.332	&46.095&	21.369&	23.225	&23.046	&19.888	&19.906	&21.672	&\textbf{18.114}	&\textbf{18.223}	&\textbf{18.088}
        \\
        &DCNV2 & 
        23.120	&45.864	&21.376	&22.718	&21.990	&19.823	&20.026	&21.245	&\textbf{18.083}	&\textbf{18.213}	&\textbf{18.068}
        \\
        &GDCN &
        23.718	&45.899&	21.388	&22.820	&22.235	&19.808	&20.693	&21.600	&\textbf{18.071}	&\textbf{18.218}	&\textbf{18.058}
        \\
        
        \hline
        &MLP & 
        0.6781&	0.5679&	0.4566&	0.6925&	0.6978&	0.6888&	0.7096&	0.7000&	\textbf{0.7137}	&\textbf{0.7151	}&\textbf{0.7178}
        \\
        XAUC &DCN & 
        0.6871&	0.5672&	0.3771 &	0.6880&	0.6900 &	0.6863&	0.7099&	0.7037&	\textbf{0.7134}&	\textbf{0.7160}&	\textbf{0.7181}        
        \\
        &DCNV2 &0.6845 &	0.5673 &	0.3856 &	0.6922 &	0.6978 &	0.6864	& 0.7092 &	0.7034 &	\textbf{0.7119}&	\textbf{0.7150} &	\textbf{0.7172 }
        \\
        &GDCN & 0.6712 &	0.5673 &	0.3907 &	0.6911 &	0.6964 &	0.6869 &	0.7091 &	0.7015 & \textbf{0.7140} &	\textbf{0.7153} &	\textbf{0.7181} \\
        \hline
        &MLP & 0.6521 &	0.5925 &	0.5335 &	0.6617 &	0.6641 	&0.6549 &	0.6645 	&0.6658 &	\textbf{0.6672} &	\textbf{0.6717} & \textbf{0.6725}
        \\
        XGAUC &DCN & 0.6630 & 0.5921 & 0.4633 & 0.6600 & 0.6609 & 0.6542 & 0.6641 & 0.6676 & \textbf{0.6686} & \textbf{0.6716} & \textbf{0.6735}
        \\
        &DCNV2 & 0.6624 	&0.5922 	&0.4687 	&0.6616 	&0.6632 	&0.6542 &	0.6642 &	0.6672& 	\textbf{0.6671}& 	\textbf{0.6718} & \textbf{0.6729}
        \\
        &GDCN & 0.6578& 	0.5921 	&0.4671 &	0.6612 &	0.6636 &	0.6543& 	0.6646 &	0.6669 &	\textbf{0.6690} 	&\textbf{0.6722} & \textbf{0.6739}
        \\
        \hline
    \end{tabular}  
    \caption{Raw staytime error and accuracy metrics for relative advantage debiasing methods versus baselines.}
    \label{tab:watchtime_prediction}
\end{table*}

We evaluate models using standard metrics from literature \cite{D2Q}, including: (1) mean absolute error (MAE), which measures the average absolute difference between predicted and true watch times; (2) XAUC, an extension of AUC for continuous outcomes to assess how well model predictions preserve the ranking order of engagement between item pairs; and (3) group-XAUC, which measures whether predicted scores preserve the true watch-time ordering of item pairs within each group. For online A/B tests, we track common engagement metrics such as watch time, finish rate, skip rate, etc. Extensions such as generalized group AUC for user or video cohorts, as well as user clustering procedures, are introduced and discussed in the relevant result sections. Duration binning is also employed, as described earlier in the methods. 

\subsection{Watch-Time Prediction}
Our primary hypothesis is that RAD’s bounded, homoskedastic quantile labels can reduce prediction error by stabilizing training and removing systematic biases from watch-time data. To test this, we train models with user-side (RAD-U) and video-side (RAD-V) labels across multiple backbone architectures. After training, predicted quantiles are mapped back to the watch-time domain using empirical inverse CDFs estimated from the training set. Following standard practice in previous works, watch time is not truncated at video duration. Because users frequently watch videos past the initial ``finish" point of the video, untruncated watch time provides additional preference signals.

Table~\ref{tab:watchtime_prediction} reports mean absolute error (MAE), XAUC, and XGAUC across all architectures. Both RAD-U and RAD-V outperform all baselines, confirming that quantile-based labels lead to more accurate and robust watch-time prediction. Specifically, RAD achieves lower MAE, stronger ranking performance (XAUC), and superior user-centric ranking (XGAUC) compared to all other methods. These results support our hypothesis  that quantile-based, distributionally normalized labels not only improve overall accuracy, but also enhance the model’s ability to capture relative user preferences in a more stable and consistent way.

To further leverage complementary information from user and video perspectives, we average their watch-time predictions after transforming the predicted quantiles back to the watch-time domain using the empirical inverse CDF from the training data. This combined approach (RAD-UV) consistently achieves the best or near-best results for all metrics, outperforming either RAD-U or RAD-V individually. The improvement suggests that each perspective captures different aspects of user engagement, and their prediction errors offset each other. By averaging, we reduce individual model biases and variance, leading to more robust, accurate, and user-centric predictions, highlighting the value of integrating both perspectives in debiased modeling.

\begin{table*}[ht]
    \centering
    
    \small 
    
    \begin{tabular}{c|c|cccccccc|cc|c}
    \hline
        Metric & Backbone & VR & PCR & WLR & WTG &D2Co & D2Q &CWM & CQE &\begin{tabular}[c]{@{}c@{}}RAD-V\\ CDF\end{tabular}&\begin{tabular}[c]{@{}c@{}}RAD-U\\ CDF\end{tabular} & \begin{tabular}[c]{@{}c@{}}RAD-UV\\ CDF\end{tabular}\\
        \hline
        &MLP & 0.6909 & 0.6613 & 0.6068 & 0.7047 & 0.7054 & 0.6984 & 0.7050 & 0.7050 & --- & \textbf{0.7105} & \textbf{0.7127} \\
        
        UGroup & DCN & 0.7017 & 0.6603 & 0.5463 & 0.7044 & 0.7043 & 0.6971 & 0.7065 & 0.7067 & --- & \textbf{0.7092} & \textbf{0.7132} \\

        XAUC & DCNV2 & 0.6994 & 0.6607 & 0.5551 & 0.7048 & 0.7044 & 0.6975 & 0.7068 & 0.706 & --- & \textbf{0.7105} & \textbf{0.7124} \\

        &GDCN & 0.6954 & 0.6603 & 0.5521 & 0.7053 & 0.7052 & 0.6971 & 0.7055 & 0.7057 & --- & \textbf{0.7101} & \textbf{0.7128} \\

        \hline
        
        &MLP & 0.6389 	&0.6748 &	0.5172 	&0.6752 	&0.6739 	&0.6755 &	0.6713 	&0.6728 	&\textbf{0.6803} 	&---	& \textbf{0.6786} \\
        VGroup & DCN & 0.6599 &	0.6751 	&0.4592 &	0.6725 &	0.6722 &	0.6747 &	0.6716 	&0.6742 &	\textbf{0.6793} 	&---	& \textbf{0.6783} \\
        XAUC &DCNV2 & 0.6533 &	0.6749 &	0.4749 &	0.6754 	&0.6749 &	0.6745 	&0.6716 	&0.6736 &	\textbf{0.6785} &	---	& \textbf{0.6776} \\
        &GDCN & 0.6306 &	0.6753 &	0.4853 &	0.6749 &	0.6747 &	0.6743 &	0.6700 &	0.6728 &	\textbf{0.6805} & --- & \textbf{0.6771} \\
        \hline
    \end{tabular}
    \caption{User and video grouped XAUC metrics. In relative advantage debiasing methods, XAUC is calculated directly based on CDF values. The combined RAD method, RAD-UV, estimates a joint CDF computed via Bayesian evidence fusion.}
    \label{tab:xauc_cdf}
\end{table*}

\subsection{Relative Preference Modeling}
While previous sections focused on reconstructing raw watch time from debiased RAD predictions, our broader goal is to assess how well the proposed methods capture underlying user preferences—that is, whether debiased labels yield more accurate orderings within each user and within each video. To this end, we model RAD labels directly and evaluate whether they preserve the ordering of watch time across both axes. This formulation enables a focused assessment of each model’s ability to capture latent preference structures using RAD labels, independent of engagement magnitude.
\subsubsection{Group-level Metrics}
To assess each model’s ability to capture user–video preferences, we generalize the standard XAUC metric into two group-level variants tailored for evaluating our quantile-based approach.

The original XGAUC evaluates, for each user, how well the predicted watch-time order matches the true watch-time order, after mapping RAD predictions back to the watch-time scale. This score is then averaged across users. In our framework, we directly compare the orderings given by RAD predictions in the quantile space to the actual watch-time order, focusing on scale-free preference modeling.

\begin{itemize}
\item \textbf{User Group XAUC}: For each user, measures how well the RAD-V prediction's order matches the true watch-time order across videos.
\item \textbf{Video Group XAUC}: For each video, measures how well RAD-U predictions can rank users in a consistent way compared to their actual watch times.
\end{itemize}

These group-based metrics provide a more robust and interpretable evaluation of preference modeling by focusing directly on orderings in the quantile space.

\subsubsection{Relative Preference Prediction}
Table~\ref{tab:xauc_cdf} reports XAUC scores computed directly on model-predicted CDF values. Across models, debiasing-based approaches—including D2Q, CWM, and all RAD variants—consistently outperform classical baselines such as VR, PCR, and WLR. This highlights the importance of explicitly correcting for confounding effects in modeling user engagement and preference. Among all debiasing methods, RAD achieves the strongest overall performance: RAD-U achieves the highest scores on User Group XAUC, while RAD-V leads on Video Group XAUC, supporting our hypothesis that side-specific debiasing improves intra-group preference modeling.

Notably, the fused RAD-UV model, which combines user- and video-side predictions in the latent space, matches or exceeds the performance of the single-sided models and all baselines on group XAUC (Table~\ref{tab:xauc_cdf}). This demonstrates a clear benefit of jointly modeling both sides for a balanced and robust performance. Overall, these findings demonstrate that quantile-based debiasing with RAD, particularly when integrating different umbrella factors, yields the most accurate and robust preference rankings.

\setlength{\textfloatsep}{6pt}
\begin{figure*}[ht]
    \centering
    
    \includegraphics[width=1\linewidth]{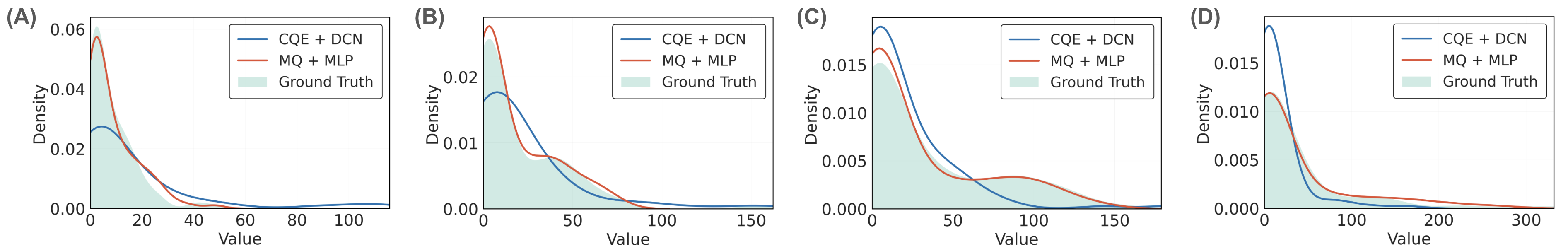}
    \caption{Gaussian kernel density estimates of predicted versus ground‐truth watch‐time distributions for user‐side clusters grouped by training‐set size quartile: (A) bottom 25 \%, (B) 25–50 \%, (C) 50–75 \%, and (D) top 25 \%. CQE’s single‐stage estimates (blue) fail to capture the lower‐value regions in (A) and (D), miss secondary modes in (B) and (C), and underestimate heavy tails in (A), (B), and (D), leading to larger errors. In contrast, the MQ + MLP multiquantile model (red) more closely follows the ground-truth curves (green) across all cohort sizes, effectively modeling both typical and irregular shapes.}
    \label{fig:distib_comparisons}
\end{figure*}

\input{LaTeX/tables/distribution_comparisons}

\setlength{\textfloatsep}{6pt}
\begin{table}[t]
    \centering
    \begin{tabular}{c|ccc}
\hline
\textbf{XGAUC} & \textbf{CQE} & \textbf{MQ + RAD-U} & \textbf{RAD-U} \\ \hline
MLP & 0.7044 & 0.7101 & 0.7131 \\
DCN & 0.7071 & 0.7105 & 0.7136 \\
DCN\_V2 & 0.7066 & 0.7099 & 0.7128 \\
GDCN & 0.7059 & 0.7100 & 0.7130 \\ \hline
\end{tabular}
    \caption{RAD with multiquantile estimation of the user-side distribution (RAD-U + MQ) consistently outperforms CQE, and closely matches performance of RAD-U with exact CDF labels. 
    }
    \label{tab:comparison_final}
\end{table}

\subsection{Two-Stage Architecture Ablations}
RAD uses umbrella factors to debias a wide range of confounding signals, outperforming single-factor methods such as D2Q. Interestingly, RAD also surpasses CQE, although both leverage watch-time distributions for preference learning. The key distinction is that RAD explicitly separates distribution estimation from preference modeling, whereas CQE entangles both in a single joint process. In this ablation, we address two key objectives: (1) evaluating whether RAD’s two-stage design offers advantages over the one-stage CQE architecture, and (2) testing whether costly per-ID empirical quantiles can be replaced by a compact, learnable distributional embedding without sacrificing accuracy or overall model performance.

\subsubsection{Learnable Distribution Embedding}
In online deployment, converting watch times to RAD labels typically requires storing and querying large watch-time histories, which adds overhead from additional indexing services and increases latency. To address this, we propose a distributional embedding approach that learns the distributional information directly as neural network parameters, thereby eliminating the need for historical data retrieval. Each umbrella factor (e.g., user or video cohort) is mapped to a learned embedding, which is then processed by a shared MLP to produce $K=100$ raw logits. After a ReLU activation and cumulative sum, these logits form a non-negative, strictly increasing sequence of quantile estimates.
The model is trained using quantile regression to align the predicted quantiles with empirical watch-time quantiles. At inference, this setup enables the quantile function to be generated without external lookup. In our ablation studies, we systematically compare the accuracy of this learnable embedding approach with baseline methods, including empirical quantile estimation and CQE.

\subsubsection{Mitigating Sparsity with User Clustering}
Accurately learning user- or video-side distributional information through quantile embeddings requires robust and statistically supported quantile estimates. However, in the KuaiRand dataset, neither individual user nor video cohorts offer enough samples for stable quantile learning. For example, the lowest 10th percentile of user and video cohorts—ranked by sample size—contains no more than ten samples per cohort (see Appendix), raising concerns about the reliability of quantile embedding learning and its evaluation. To mitigate this potential issue in our ablation analyses, we cluster users into ten groups using K-modes clustering~\cite{k_modes} on sparse user features (see Appendix), treating each group as a clustered analogue of the user ID. This approach helps ensure each cohort has sufficient samples for robust quantile estimation and embedding training. In our ablation studies, these cluster IDs serve as the umbrella condition for deriving user-side RAD labels.

\setlength{\textfloatsep}{6pt}
\begin{table}[t]
    \centering
    \resizebox{0.47\textwidth}{!}{
    \begin{tabular}{c|cccc}
    \hline 
        \textbf{Methods} & \textbf{CQE}& \textbf{RAD-U}  & \textbf{RAD-V} &  \textbf{RAD-UV}  \\
    \hline
        User Group XAUC & 0.623  & 0.642 & ---    & 0.637 \\
        Video Group XAUC & 0.594  & ---  & 0.631   & 0.625 \\
    \hline
    \end{tabular}
    }
    \caption{Group XAUC for offline industrial dataset.}
\label{tab:some_table_for_rad_cqe_offline_ind}
\end{table}

\input{LaTeX/tables/online_results}
\subsubsection{Two-stage Benefits and Embedding Efficiency}
To evaluate the two-stage architecture and distributional embeddings, we compare (1) RAD-U with empirical quantiles (two-stage), (2) RAD-U with learnable distribution embeddings (two-stage + embeddings), and (3) CQE (one-stage).

We first assess distribution matching quality using the 1-Wasserstein distance (Table~\ref{tab:dist_compare}), where the empirical-quantile baseline (no learning) sets the minimum achievable error between training and test sets. For CQE, predicted user–video distributions are aggregated into a user-side cohort distribution via Monte Carlo mixture sampling~\cite{neal1992bayesian}. RAD-U’s learned embeddings closely approach this lower bound and outperform CQE, demonstrating that separating CDF estimation from preference learning produces more accurate distribution fits without requiring storage of watch-time logs. Figure~\ref{fig:distib_comparisons} provides representative examples, showing that RAD-U’s multiquantile embeddings consistently align with the ground truth for various user group sizes, while CQE’s estimates diverge.

To evaluate preference modeling, we compare methods using User Group XAUC (Table~\ref{tab:comparison_final}). Across all architectures, both RAD-based approaches, either with empirical quantiles or learned distribution embeddings, consistently outperform the one-stage CQE method. This further underscores the advantage of the two-stage design observed in distribution matching. We also report MAE and XAUC for watch-time prediction, where both two-stage methods consistently surpass CQE, with the embedding-based approach achieving results close to those of empirical quantiles. Taken together, these findings highlight the robustness and accuracy of the two-stage architecture. At the same time, distributional embeddings deliver near equivalent predictive performance while providing greater efficiency and scalability, making them more practical for real-world deployment.
\subsubsection{Offline Industrial Dataset}
We also evaluate RAD and CQE on a large-scale offline industrial dataset. In this setting, RAD employs distributional embeddings to estimate quantiles for real-world systems. Models are trained on 14 days of data and evaluated on a held-out day. With ample samples per umbrella factor, group XAUC is computed directly for both user-side (RAD-U) and video-side (RAD-V) without clustering. As shown in Table~\ref{tab:some_table_for_rad_cqe_offline_ind}, RAD consistently and substantially outperforms CQE on both user-group and video-group XAUC, validating its robustness in high-data scenarios for industrial recommendation.

\subsection{Online A/B Experiments}

We further evaluate the RAD methods in a large-scale short-video recommendation platform. In this setting, the abundance of data allows us to test robustness and effectiveness in high-data regimes and real-world settings.
\paragraph{Implementation Details} We directly learn  and deploy three RAD-based models, RAD-U, RAD-V, and their Bayesian fusion (RAD-UV), as candidate rankers for user–item preference. The current production model serves as the baseline. For online evaluation, we compare the outcomes of these three new models against the baseline, forming four experimental groups. Each group is assigned an equal portion of platform traffic, ensuring fair comparison under consistent conditions.
\paragraph{Results}
Table~\ref{tab:online_results} reports the results of large-scale online experiments. RAD-U produces the largest gain in Finish Playing, while RAD-V achieves the highest result for Watch Count, reflecting the complementary advantages of user- and video-side debiasing. All RAD-based methods improve user engagement over the baseline, with the combined RAD-UV model showing the most balanced gains, leading in Active Days, Watch Time, and Skip Rate. These findings highlight both the overall effectiveness of RAD-based debiasing and the added value of integrating user- and video-side signals for real-world recommendation.

\section{Conclusion}

We have presented Relative Advantage Debiasing (RAD), a principled framework for mitigating confounding effects in watch-time-based recommendation. By reframing raw watch time as a cohort-relative quantile, RAD systematically and jointly corrects both user- and item-side biases through the utilization and integration of umbrella factors. Our learnable distributional embedding enables efficient, scalable deployment without the need for historical data storage. Both offline and online experiments demonstrate that RAD consistently improves watch-time prediction and ranking quality over existing baselines, with the hybrid RAD showing most balanced gains across key engagement metrics.

RAD opens several promising directions for the broader recommendation and learning community. For example, RAD quantiles make it possible to jointly optimize watch time with various interaction signals (such as likes and comments) within a normalized space, thus providing a natural foundation for multi-task learning \cite{lu2018like}. In the context of reinforcement learning, quantile-based rewards can help stabilize policy optimization and readily connect to reward transformation frameworks such as Group Relative Policy Optimization (GRPO) \cite{shao2024deepseekmath}. These normalized signals may also enhance generative listwise recommendation \cite{liu2023generative}, supplying calibrated feedback for more effective credit assignment in sequence or slate-based models \cite{bello2018seq2slate}. Moreover, since quantile labels inherently adjust for duration and popularity, future work should investigate their effects on fairness, robustness, and cold-start performance \cite{zhu2021fairness}, as well as their applicability to other domains—including music, news, and e-commerce \cite{chen2024novel}.

\section{Acknowledgments}
We appreciate Kunpeng Liu and Danhui Guan for foundational work and helpful discussions that refined the method.






\nocite{10184833}
\nocite{cade2003}
\nocite{10.5555/3504035.3504388}
\nocite{dorka2024}
\nocite{10.1145/582415.582418}
\nocite{10.1145/3643888}
\nocite{10.1145/3308558.3313513}
\nocite{10.1145/3459637.3482417}
\nocite{7780901}
\nocite{PETNEHAZI2021100096}
\nocite{10.1145/3539597.3570374}
\nocite{10.1145/3617826}
\nocite{8985289}
\nocite{tang22}
\nocite{swat}
\nocite{10.1145/3583780.3615483}
\nocite{10.1145/3097983.3098134}
\bibliography{aaai2026}

\include{LaTeX/appendix}

\end{document}

%% file: LaTeX/tables/distribution_comparisons.tex
\setlength{\textfloatsep}{5pt}

\begin{table}[ht]
    \centering

    \begin{tabular}{l|c}
        \hline
        \textbf{Method} & \textbf{Wasserstein Distance}\\
        \hline
        CQE + MLP & 9.3945\\
        CQE + DCN & 9.0804 \\
        CQE + DCN-V2& 9.2101 \\
        CQE + GDCN & 9.1529 \\
        MQ + MLP & 2.2793  \\
        Minimum Wasserstein Distance & 2.1500 \\
        \hline
    \end{tabular}
    
    \caption{Average Wasserstein distance per estimated user group distribution, comparing CQE with MLP based multiquantile estimation.}
    \label{tab:dist_compare}
\end{table}

%% file: LaTeX/tables/online_results.tex
\setlength{\textfloatsep}{6pt}
\begin{table}[ht!]
\centering
\resizebox{0.47\textwidth}{!}{
\begin{tabular}{l|ccccc}
\hline
       & \textbf{Active Days} & \textbf{Watch Time} & \textbf{Watch Count} & \textbf{Finish Playing} & \textbf{Skip Rate}      \\ \hline
RAD-U  & 0.0225\%    & 0.2530\%   & 0.0451\%    & 1.3296\%       & -0.4347\% \\
RAD-V  & 0.0160\%    & 0.0562\%   & 0.2554\%    & 0.7531\%       & -0.1887\% \\
RAD-UV & 0.0290\%    & 0.3246\%   & 0.0744\%    & 1.1125\%       & -0.864\% \\ \hline
\end{tabular}
}
\caption{Metrics for online experiments with relative advantage debiasing.}
\label{tab:online_results}
\end{table}

%% file: LaTeX/appendix.tex
\section{Appendix}

\subsection{Proof of Proposition~\ref{prop:variance-monotonicity}: Variance monotonicity under umbrella sufficiency}
\label{app:proof_quantile_uniformity}\label{app:proof_var_monotonicity}
Since $\sigma(c^{(k)}) \subseteq \sigma(i)$, the tower property of conditional expectation gives
\begin{equation}
  \mathbb{E}[S_{u,i} \mid c^{(k)}]
  = \mathbb{E}\bigl(\mathbb{E}[S_{u,i}\mid i]\mid c^{(k)}\bigr).
\end{equation}
Applying the law of total variance to $\mathbb{E}[S_{u,i}\mid i]$ with respect to $c^{(k)}$ yields
\begin{align}
  \mathrm{Var}\bigl(\mathbb{E}[S_{u,i}\mid i]\bigr)
  &= \mathrm{Var}\bigl(\mathbb{E}[S_{u,i}\mid c^{(k)}]\bigr) \notag \\
  &\quad + \mathbb{E}\Bigl[\mathrm{Var}\bigl(\mathbb{E}[S_{u,i}\mid i]\mid c^{(k)}\bigr)\Bigr] \notag \\
  &\ge \mathrm{Var}\bigl(\mathbb{E}[S_{u,i}\mid c^{(k)}]\bigr)
\end{align}
which proves the proposition on variance monotonicity in the main text.
%
%
%

\subsection{Proof of Proposition~\ref{prop:quantile_uniformity}: Conditional Uniformity and Independence of Quantile Labels}
\label{app:proof_quantile_uniformity}

By the probability integral transform, for any realization $G_k$ of $G$, the quantile label $Q_{u,i}(G_k) = F_{S|G=G_k}(S_{u,i})$ satisfies
\begin{equation}
\Pr\left( Q_{u,i}(G_k) \leq q \mid G = G_k \right) = q, \quad \forall\, q \in [0,1].
\end{equation}
Thus, $Q_{u,i}(G_k) \mid G = G_k \sim \mathrm{Uniform}(0,1)$, which always has mean $1/2$ and variance $1/12$ regardless of $k$.

For the marginal (unconditional) distribution, considering $G$ as a random variable, we have
\begin{equation}
\Pr\left( Q_{u,i}(G) \leq q \right) = \mathbb{E}_G\left[ \Pr\left( Q_{u,i}(G) \leq q \mid G \right) \right] = q.
\end{equation}
Therefore, $Q_{u,i}(G) \sim \mathrm{Uniform}(0,1)$, and since this marginal distribution does not depend on $G$, it follows that $Q_{u,i}(G)$ is statistically independent of $G$.



\subsection{Baseline Method Descriptions}
The benchmark methods we use for comparison in this paper are described in greater detail below.

\begin{itemize}
    \item \emph{Value Regression (VR)}: The VR baseline directly predicts video watch time as a continuous scalar value. Value regression is a debiasing-free method, and provides a reference point for all debiasing methods.
    \item \emph{Play Completion Rate (PCR)}: PCR normalizes watch time by dividing it by the total video duration, producing a proportion of video watched. This normalizes for length and reduces duration bias.
    \item \emph{Weighted Logistic Regression (WLR)}: WLR treats user engagement as a binary classification problem (e.g., whether watch time exceeds a fixed threshold), and assigns weights to training examples based on estimated propensities.
    \item \emph{Watchtime Gain (WTG)}: WTG estimates how much a video’s watch time exceeds a reference baseline, such as the average engagement level of the user. This method captures relative user engagement, centering predictions on individualized expectations.
    \item \emph{Debiased and Denoised watch time Correction (D2Co)}: D2Co applies inverse propensity weighting to adjust for exposure bias and denoising techniques to reduce variance. It treats debiasing as a two-step process: first correcting for selection effects, then smoothing noisy feedback signals.
    \item \emph{Duration-debiased Quantiles (D2Q)}: D2Q predicts the quantile position of a watch time value within a duration-specific distribution (duration bin). This reduces the bias introduced by differing video lengths, allowing for more robust comparisons across heterogeneous content. 
    \item \emph{Counterfactual Watch Model (CWM)}: The CWM uses counterfactual inference to estimate watch time as if the video were presented under standardized conditions (e.g., fixed position, slot, or exposure context). This approach removes contextual confounding by simulating a controlled environment.
    \item \emph{Conditional Quantile Estimation (CQE)}: CQE estimates the conditional distribution of watch time using quantile regression, conditioned on both user and video features. It outputs calibrated quantile predictions, capturing uncertainty and variability in user preferences.

\end{itemize}

\subsubsection{Cohort Sample Counts by User and Video}
Table~\ref{tab:conditional_distribution_counts_single} reports sample counts per cohort by percentile for user-side and video-side groupings. For each percentile, the table reports the minimum number of samples present in a cohort at or above that percentile; for example, at the 10th percentile, a user-side cohort contains 8 samples and a video-side cohort contains 9. Both user-side and video-side cohorts show potential concerns regarding small sample sizes at the lowest percentiles, with the issue being more pronounced on the user side, which may further affect the reliability of quantile estimation for certain cohorts.


\setlength{\textfloatsep}{6pt}
\begin{table}[!ht]
  \centering
  \setlength{\tabcolsep}{5pt}           

  \begin{tabular}{c|cc||c|cc}
    \hline
    Percentile & User & Video & Percentile & User & Video \\ \hline
    10 &   8  &     9 &  60 &   49 &     90 \\
    20 &  15  &    16 &  70 &   62 &    143 \\
    30 &  21  &    26 &  80 &   81 &    242 \\
    40 &  29  &    38 &  90 &  112 &    490 \\
    50 &  39  &    57 & 100 &  895 &  10376 \\ \hline
  \end{tabular}

  \caption{ Sample counts per cohort by percentile for user-side and video-side groupings. Left block: 10th–50th percentiles; right block: 60th–100th percentiles.}
  \label{tab:conditional_distribution_counts_single}
\end{table}

\subsubsection{Hyperparameters}
Following an established protocol~\cite{CWM}, all backbone models use three hidden layers of dimension size 64. Model training employed a learning rate of 1e-5, the Adam optimizer, a maximum of 50 epochs, and early stopping with a patience of 5 epochs, as validation consistently showed convergence within this range. CWM models were optimized using the log counterfactual likelihood over observed watch times. For CQE and for the first stage of RAD (when using distribution embedding for quantile estimation), we employed the pinball loss for quantile prediction, while the second stage models of RAD were trained using mean squared error (MSE) loss.

\subsection{User Features for K-Modes Clustering}

K-Modes clustering is performed using the following as user features: Number of fans, number of friends, number of accounts the user follows, user activity level, number of days since registration, and 18 encoded categorical features provided by the KuaiRand dataset. A complete list of features is given in the KuaiRand dataset description. All numeric features are first grouped into categorical data in the form of ranges.
\subsection{Sensitivity of Bayesian Fusion to $\beta/\alpha$}
\textcolor{black}{We explore a variety of $\beta/\alpha$, as shown in Table \ref{tab:fusion_sensitivity}, and find that Bayesian fusion is not sensitive to a wide range of ratios. Larger ratios favor video-side predictions, as more weight is placed on the video-side CDF, and similarly, smaller ratios favor user-side predictions. Using precision-weighted user/video counts as a proxy for $\alpha$ and $\beta$ establishes a balanced tradeoff between video and user side signals.}
\begin{table}[]
    \centering
  \setlength{\tabcolsep}{5pt}           

  \begin{tabular}{c|cc}
    \hline
    $\beta/\alpha$ & User Group XAUC & Video Group XAUC \\ \hline
    Baseline & 0.7057 &
    0.6749 \\
    \hline
    0.1 & 0.7092 & 0.6784 \\
    0.2 & 0.7112 & 0.6778 \\
    0.5 & 0.7124 & 0.6773 \\
    1.0 & 0.7128 & 0.6771 \\
    2.0 & 0.7129 & 0.6770 \\
    5.0 & 0.7130 & 0.6769 \\
    10. & 0.7130 & 0.6768 \\
    \hline
  \end{tabular}
    \caption{Using a GDCN backbone model, effect of $\beta/\alpha$ on results for RAD-UV in user- and group-side XAUC, compared to the highest baseline values. A higher ratio favors video-side predictions and a lower ratio favors user-side predictions. The results suggest that RAD-UV is robust to $\beta/\alpha$. }
    \label{tab:fusion_sensitivity}
\end{table}

\subsection{RAD Hyperparameter Sensitivity}

\begin{table}[]
    \centering
  \setlength{\tabcolsep}{5pt}           
\begin{tabular}{c|c|ccc}
\hline
Hyperparameter & Setting & MAE & XAUC & XGAUC \\ \hline
User Cluster & K=5 & 18.218 & 0.7153 & 0.6722 \\
Count & K=10 & 18.216 & 0.7152 & 0.6720 \\
 & K=20 & 18.210 & 0.7155 & 0.6722 \\ \hline
Multiquantile & Q=100 & 18.989 & 0.7054 & 0.6641 \\
Dimension & Q=200 & 18.985 & 0.7052 & 0.6642 \\
 & Q=500 & 18.992 & 0.7057 & 0.6642 \\ \hline
User Duration & D=2 & 18.218 & 0.7153 & 0.6722 \\
Bin Count & D=4 & 18.331 & 0.7136 & 0.6730 \\
 & D=8 & 18.169 & 0.7152 & 0.6714 \\ \hline
\end{tabular}
\caption{\textcolor{black}{Sensitivity of RAD results to various method-specific hyperparameters. RAD shows stable results and outperforms baselines across a wide range of hyperparameter choices. Results reported on GDCN model.}}

\label{tab:rad_u_hypers}
\end{table}
\textcolor{black}{We investigate the effect of \emph{K-Modes cluster count}, \emph{Multiquantile embedding dimension}, and \emph{Duration-bin count} on user-side debiasing. As shown in Table \ref{tab:rad_u_hypers}, results are not sensitive to different hypereparameter values.}

\subsection{Pseudo-code for RAD Label Estimation}

Below are two algorithms for computing RAD quantile labels from raw watch-time data, both mapping raw watch times ${S_{u,i}}$ to normalized quantile scores ${Q_{u,i}}\in(0,1)$. The first assigns cohort-relative quantiles via empirical CDF lookup, while the second uses a learnable embedding model to estimate quantile breakpoints.

\FloatBarrier

\begin{algorithm}[H]
\caption{Empirical CDF–Based RAD Labels}
\begin{algorithmic}[1]
  \STATE \textbf{Input:} Watch-time records $\{(u, v, S_{u,v})\}$; cohort assignment $c$ for each record
  \STATE \textbf{Output:} Quantile labels $\{Q_{u,v}\}\subset(0,1)$
  \FOR{each cohort $c$}
    \STATE $\mathcal{S}_c \leftarrow \{S_{u,v} \mid (u,v) \text{ in } c\}$
    \STATE $N_c \leftarrow |\mathcal{S}_c|$
    \STATE $(S^{(1)}_c \leq S^{(2)}_c \leq \cdots \leq S^{(N_c)}_c) \leftarrow \operatorname{sort}(\mathcal{S}_c)$
    \FOR{each $(u, v)$ in $c$}
      \STATE $r_{u,v} \leftarrow |\{j : S^{(j)}_c \le S_{u,v}\}|$
      \STATE $Q_{u,v} \leftarrow r_{u,v} / N_c$
    \ENDFOR
  \ENDFOR
\end{algorithmic}
\end{algorithm}

\FloatBarrier
\begin{algorithm}[H]
\caption{Learnable Quantiles for RAD Labels}
\begin{algorithmic}[1]
  \STATE \textbf{Input:} Watch-time records $\{(u, v, S_{u,v})\}$; cohort assignment $c$ for each record; embedding table $E$; shared MLP $f_\psi$
  \STATE \textbf{Output:} Quantile labels $\{Q_{u,v}\}\subset(0,1)$
  \FOR{each cohort $c$}
    \STATE $e_c \leftarrow E(c)$ \COMMENT{learnable embedding for $c$}
    \STATE $(\ell_1, \ldots, \ell_K) \leftarrow f_\psi(e_c)$ \COMMENT{shared MLP outputs $K$ logits}
    \STATE $(b_1, \ldots, b_K) \leftarrow \operatorname{cumsum}(\operatorname{ReLU}(\ell_1), \ldots, \operatorname{ReLU}(\ell_K))$ \COMMENT{increasing breakpoints}
    \FOR{each $(u, v, S_{u,v})$ in $c$}
      \STATE $k \leftarrow \min\{j : S_{u,v} \leq b_j\}$
      \STATE $Q_{u,v} \leftarrow k/K$
      \COMMENT{use quantile loss on $(b_1,\ldots,b_K)$ vs.\ $S_{u,v}$ during training}
    \ENDFOR
  \ENDFOR
\end{algorithmic}
\end{algorithm}

\FloatBarrier